\documentclass{article}

\usepackage[numbers]{natbib}
\usepackage[utf8]{inputenc}
\usepackage{graphicx}
\usepackage{float}
\usepackage{subfigure}
\usepackage{url}
\usepackage{colortbl}
\usepackage{amssymb}
\usepackage{pifont}
\usepackage{float}
\usepackage{tablefootnote}
\usepackage{amsmath}
\usepackage[dvipsnames]{xcolor}
\usepackage[hidelinks]{hyperref}
\usepackage{placeins}

\hypersetup{
  colorlinks   = true,
linkcolor=NavyBlue
,citecolor=NavyBlue
,filecolor=Red
,urlcolor=NavyBlue
,menucolor=Red
,runcolor=Red
,linkbordercolor=NavyBlue
,citebordercolor=NavyBlue
,filebordercolor=Red
,urlbordercolor=NavyBlue
,menubordercolor=Red
,runbordercolor=Red
}

\def\etc{etc.}

\usepackage{arxiv}

\title{3rd Continual Learning Workshop Challenge on Egocentric Category and Instance Level Object Understanding}

\author{
Lorenzo Pellegrini\\
Department of Computer Science\\
University of Bologna\\
\And
Chenchen Zhu\\
Meta AI\\
\And
Fanyi Xiao\\
Meta AI\\
\And
Zhicheng Yan\\
Meta AI\\
\And
Antonio Carta\\
Department of Computer Science\\
Unversity of Pisa\\
\And
Matthias De Lange\\
Department of Electrical Engineering\\
KU Leuven
\And
Vincenzo Lomonaco\\
Department of Computer Science\\
Unversity of Pisa\\
\And
Roshan Sumbaly\\
Meta AI\\
\And
Pau Rodriguez\\
ServiceNow Research\\
\And 
David Vazquez\\
ServiceNow Research
}

\renewcommand{\headeright}{3rd CLVision Workshop Challenge Report}
\renewcommand{\undertitle}{3rd Continual Learning Workshop on Continual Learning in Computer Vision at CVPR Challenge Report}
\renewcommand{\shorttitle}{}

\hypersetup{
pdftitle={3rd Continual Learning Workshop Challenge on Egocentric Category and Instance Level Object Understanding},
pdfsubject={cs.CV, cs.LG, cs.AI},
pdfauthor={Lorenzo Pellegrini, Chenchen Zhu, Fanyi Xiao, Zhicheng Yan, Antonio Carta, Matthias De Lange, Vincenzo Lomonaco, Roshan Sumbaly, Pau Rodriguez, David Vazquez},
pdfkeywords={Continual Learning, Lifelong Learning, Egocentric Vision, Object Detection \& Recognition, Category-/Instance-Level Object Understanding, Competition},
}

\begin{document}
\maketitle

\begin{abstract}
    Continual Learning, also known as Lifelong or Incremental Learning, has recently gained renewed interest among the Artificial Intelligence research community. Recent research efforts have quickly led to the design of novel algorithms able to reduce the impact of the catastrophic forgetting phenomenon in deep neural networks. 
    Due to this surge of interest in the field, many competitions have been held in recent years,  as they are an excellent opportunity to stimulate research in promising directions. This paper summarizes the ideas, design choices, rules, and results of the challenge held at the 3rd Continual Learning in Computer Vision (CLVision) Workshop at CVPR 2022. The focus of this competition is the complex continual object detection task, which is still underexplored in literature compared to classification tasks. 
    The challenge is based on the challenge version of the novel EgoObjects dataset, a large-scale egocentric object dataset explicitly designed to benchmark continual learning algorithms for egocentric category-/instance-level object understanding, which covers more than 1k unique main objects and 250+ categories in around 100k video frames.
\end{abstract}


\section{Introduction}

Continual Learning (CL) is a field of Artificial Intelligence focused on studying and designing mechanisms to enable artificial systems to learn incrementally and continuously from experiences. The continual learning field presents a research perspective embracing existing research fields not only from the Computer Science and Statistics areas, such as Lifelong and Incremental Learning, but also more biologically tied ones, such as Cognitive Neuroscience. In recent years, continual learning has seen a rapid increase in popularity among AI researchers~\cite{PARISI201954}. One of the main reasons for this increase is that continual learning can be a fundamental component of many complex autonomous intelligent systems (e.g. virtual assistants, embodied robotics, autonomous driving). Apart from the theoretical point of view, the ability to learn over time is also a desirable characteristic from a practical perspective, as incremental learning allows the tailored personalization of intelligent agents.

The fundamental idea of the research on CL is that learning should happen incrementally over time given the assumption that information regarding a specific experience is available only at the time it is observed, with no (or limited) means of retrieving past data for training purposes. Because of this, the most studied problem in the continual learning field is the catastrophic forgetting phenomenon: when deep neural networks are incrementally trained using Stochastic Gradient Descent and similar means, the previously accumulated knowledge is quickly overwritten by means of abrupt parameter shifts~\cite{McCloskey1989, Robins1995}.

Among the research directions, the Computer Vision one is arguably the most explored in the continual learning literature, with many works covering benchmarks, algorithms, and metrics regarding the ability of Deep Neural Networks to learn common vision tasks such as digit and object classification \cite{GoodfellowMDCB13, Zenke2017}, reinforcement learning \cite{ring_thesis} and, more recently, object detection and segmentation \cite{inc_det, menezes_review}. The Computer Vision field is the main focus of the 3rd CLVision Workshop held at CVPR 2022, inside which the present competition was held.

\begin{figure}
    \centering
    \includegraphics[width=0.7\columnwidth]{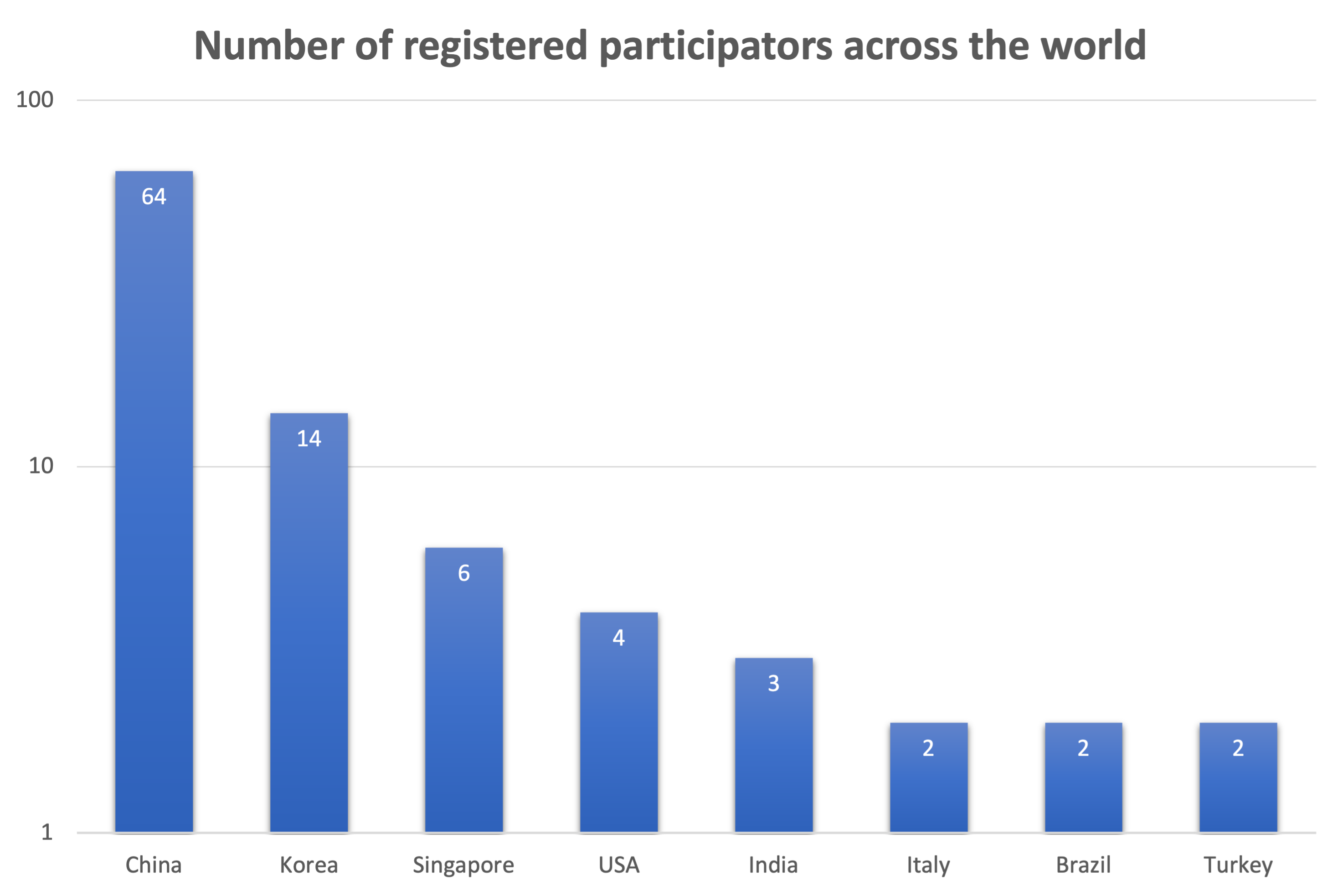}
    \caption{
        Visualization of the challenge participation across the world.
    }
    \label{fig:world_map}
\end{figure}

Several competitions have been previously organized in the context of Continual Lifelong Learning research. The focus of such competitions is either vertical, with all tracks consisting of CL problems, or mixed, with CL tasks being only a portion of the available tracks. Some Continual/Lifelong Learning competitions date back as far as 2005, such as the \textit{Pascal 2 EU challenge on covariate shift}~\cite{candela_challenge, candela_dataset_shift}. However, most CL competitions have been organized in more recent times due to the fast increase in popularity of the field. The \textit{Autonomous Lifelong Machine Learning with Drift}~\cite{escalante_automl} competition was held at NeurIPS in 2018, while the \textit{Lifelong Robotic Vision}~\cite{iros2019challenge} challenge was held at IROS in 2019. In 2021, the \textit{Self-Supervised Learning for next-generation industry-level Autonomous Driving} (SSLAD)~ \cite{sslad_challenge, sslad_dataset} competition held at ICCV included a CL track. In the same year, the \textit{Continual Semi-Supervised Learning} (CSSL)~\cite{cssl_competition} competition was held in the context of the International Workshop on Continual Semi-Supervised Learning at IJCAI 2021. The latter two included tracks consisting of i) simple detection tasks (SSLAD); ii) crowd counting tasks (CSSL). In particular, the SSLAD competition is the challenge that shares the most similarities with the competition described in this work. While detection tracks were proposed in SSLAD, the continual detection task was restricted to a limited number (6) of macro-categories of vehicles \cite{sslad_dataset}. This limited number of classes may not be enough to allow for forgetting problems of different elements of a detection model to emerge (especially the "objectness" drift, as discussed in Section \ref{sec:object_detection} and in \cite{menezes_review}).

Competitions have been organized in the context of each of the previous editions of the \textit{Workshop on Continual Learning in Computer Vision} (CLVision) held at CVPR. Each of these competitions featured novel elements of complexity that the participants had to overcome. The challenge hosted in the first edition (2020)~\cite{1st_clvision_challenge} proposed a difficult benchmark made of nearly 400 incremental experiences. The challenge hosted in the second edition (2021)~\cite{2nd_clvision_challenge} proposed a complex continual reinforcement learning benchmark.
In fact, the goal of a challenge is to stimulate the research community to produce new and more effective solutions in promising research directions. For instance, the continual learning and computer vision research fields form a long-standing duo. However, to date, most of the research efforts have been directed toward tackling the object classification problem. The importance of this problem cannot be argued, as it serves as the initial step towards building continuously learning systems for vision applications. On the other hand, the continual object detection problem (w.r.t. benchmarks, algorithms, metrics, \etc) is less explored in the literature. 
The goal of this third CLVision challenge is to push the research on continual learning for computer vision toward the detection problem. In particular, two elements of complexity were considered when designing the competition tracks: object detection and instance-level recognition.

\subsection{Object detection}\label{sec:object_detection}
Object detection is a widely studied area of the computer vision field \cite{Liu_detection_survey, Zhao_detection_review}. The ability to determine the position and identity of multiple objects in a visual input (may it consist of still images or a visual stream), is a natural ability of humans and many animals. The detection field has been widely explored and research efforts in this area are still very lively. Research does not only cover the advancement of performance under classic learning paradigms (supervised, offline training) \cite{faster-rcnn, yolo, ssd, fpn, retinanet, mask-rcnn, fcos, detr}, but also encompasses areas such as weakly-supervised learning \cite{wsddn, oicr, wildcat, wetectron}, few-shot learning \cite{fsod, metadet, meta-rcnn, tfa, mpsr, srr-fsd, defrcn}, and semi-supervised learning \cite{csd-rfcn, soft-teacher, unbiased-teacher}. It must be noted that this field is not entirely unexplored from a continual learning perspective \cite{verwimp2022re, inc_det, hayes_rodeo}. However, mainstream computer vision datasets are used (such as COCO~\cite{coco_dataset} and VOC~\cite{Everingham10}), which were not designed with continual learning in mind. In addition, most used learning techniques are already well known in the classification literature (mainly distillation and replay) with only a limited focus being put on detection-specific approaches. A review covering these initial research efforts on continual learning in object detection tasks has been recently published \cite{menezes_review}. For this competition, object detection has been chosen as the characterizing element because of its complexity and practical relevance.


Object detection is intrinsically more complex than object classification: multiple objects need to be identified in each image, objects appear at different distances and sizes, elements may be crowded and partially overlapping, etcetera. When it comes to the union of continual learning and object detection, an additional element of complexity should be considered: the ability to correctly learn to identify the position of objects. This issue is particularly evident in two-stage detectors such as Faster-RCNN, in which a specific component (the Region Proposal Network) is in charge of selecting areas of the image in which an object may be present. 
Areas of interest are selected by estimating an objectness score and then applying mechanisms such as thresholding and top-k selection. This means that, in a continual learning setting, the model not only needs to avoid catastrophic forgetting to correctly classify object classes, but it also needs to properly retain the necessary information to perform accurate localization.

For instance, let us consider the class-incremental scenario, which is arguably the most studied scenario in Continual Object Classification literature. In this setup, each training experience contains all samples from a subset of classes, and training samples for those classes are never encountered again in future experiences. This setup can be translated to object detection as follows: images are accompanied by annotations (bounding boxes and labels) related to a subset of classes only. The result is that, when incrementally training on a certain experience, the detector is trained to apply low objectness scores to areas of the image depicting objects of classes linked to other experiences. In other words, if no precautions or CL-specific techniques are used, the detector is trained to forget how to localize previous classes. For already encountered classes, the result is that the recall for those classes will drop in a possibly unrecoverable way. The same reasoning can be applied not only to the RPN but also to the classification head, which is also trained to consider these areas as background.

When designing this challenge, we kept into consideration these elements of complexity and the fact that the existing research on continual object detection is still limited. Because of this, the challenge tries to balance the push on novelty aspects and the current state of continual learning literature. While more complex mainstream datasets such as COCO~\cite{coco_dataset} and VOC~\cite{Everingham10} could be used, the complexity of those datasets would be ill-matched with the current state of the research for continual detection tasks.

Object detection is very relevant when it comes to the practical applicability of computer vision applications, especially on self-driving vehicles, virtual assistants, etcetera. Continual learning capabilities could allow for the personalization of the behavior of the intelligent system, making it more efficient and user-tailored as time passes. In particular, the dataset proposed for the competition features \emph{egocentric-view} images taken in house and work environments, which can be seen as the natural starting environment for the incremental personalization of assistants and similar intelligent agents.


\subsection{Instance-level detection and classification}
With instance recognition here we denote the ability to distinguish between specific objects. This is opposed to category-level recognition where the granularity is more coarse, and the class labels applied to objects usually follow a high-level semantic categorization. In fact, most of the object classification and detection literature focuses on category-level recognition and mainstream datasets (CIFAR~\cite{krizhevsky2009learning}, ImageNet~\cite{imagenet_dataset}, COCO~\cite{coco_dataset}, LVIS~\cite{lvis_dataset}, VOC~\cite{Everingham10}, \etc) follow this idea. On the contrary, in instance-level recognition tasks the vision system must be able to learn to discern between objects (including, for detection tasks, pinpointing their location) i) belonging to the same semantic category, and ii) sharing many visual features (which can be often related to the fact they belong to “nearby” semantic categories). Introducing recognition tasks featuring this kind of granularity certainly increases the overall complexity of the problem at hand, but it also allows the creation of more versatile artificial vision systems. Apart from objects, datasets for instance-level recognition exist for places \cite{Weyand_2020_CVPR}, art \cite{ypsilantis2021the}, and clothing \cite{liu_deepfashion}. These datasets are commonly used in the context of retrieval tasks. When it comes to continual learning, datasets featuring such granularity are CORe50~\cite{pmlr-v78-lomonaco17a} and OpenLORIS~\cite{openloris_dataset}. These continual learning datasets also share the fact that videos are provided instead of unrelated images. However, these datasets are limited in size and in the available vision tasks (classification, SLAM).


\section{EgoObjects - Challenge Version}

\begin{figure}
    \centering
    \includegraphics[width=\columnwidth]{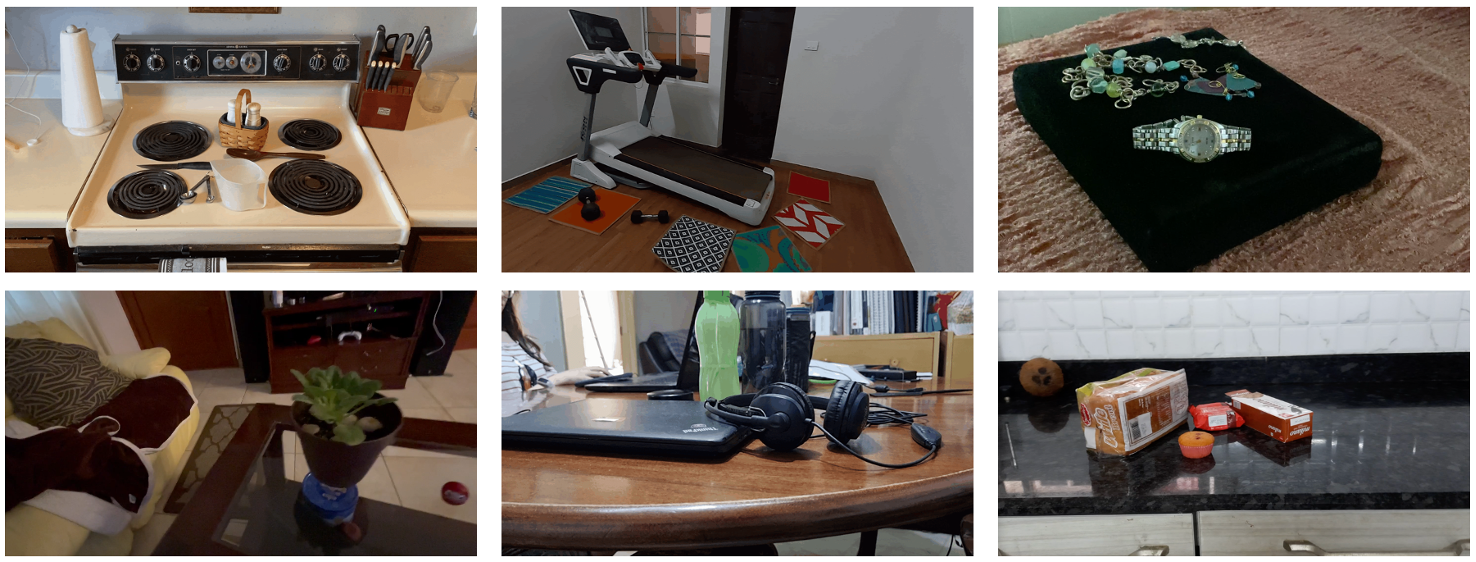}
    \caption{EgoObjects videos feature a great variety of challenging conditions in the realistic egocentric view, including distance, camera motion, background complexity and lighting.}
    \label{fig:egoobjects_intro}
\end{figure}

For this year's challenge, we aim at moving continual learning towards object-level detection and recognition from an \emph{egocentric perspective}. We also want to study the capability of incrementally recognizing specific objects, a more fine-grained setting than category-incremental learning. To this end, we setup the challenge using a curated subset of EgoObjects dataset \cite{egoobjects}, the first large-scale egocentric dataset for objects. 

The EgoObjects dataset is built with the goal of pushing the frontier of open-world object understanding. It has several main features:
\begin{itemize}
    \item Videos have been taken with a wide range of egocentric recording devices (Rayban Stories, Snap Spectacles, and Mobile) in realistic household/office environments from 25+ countries and regions.
    \item Videos feature a great variety of lighting conditions, camera-to-object distance, camera motion, and background complexity for daily indoor objects. See Figure \ref{fig:egoobjects_intro} for a snapshot of these conditions.
    \item Video frames have been sampled and annotated with rich ground truth, including 2D bounding boxes, object category labels, and object instance IDs. 
    \item Each video depicts one main object. Moreover, in the same video, the surrounding objects are also annotated.
\end{itemize}

\begin{figure}
    \centering
    \subfigure[Bar plots showing how many main objects are featured in x images. All objects in the test set are featured in at least 19 images i.e. 19 annotations.]{\includegraphics[width=0.95\columnwidth]{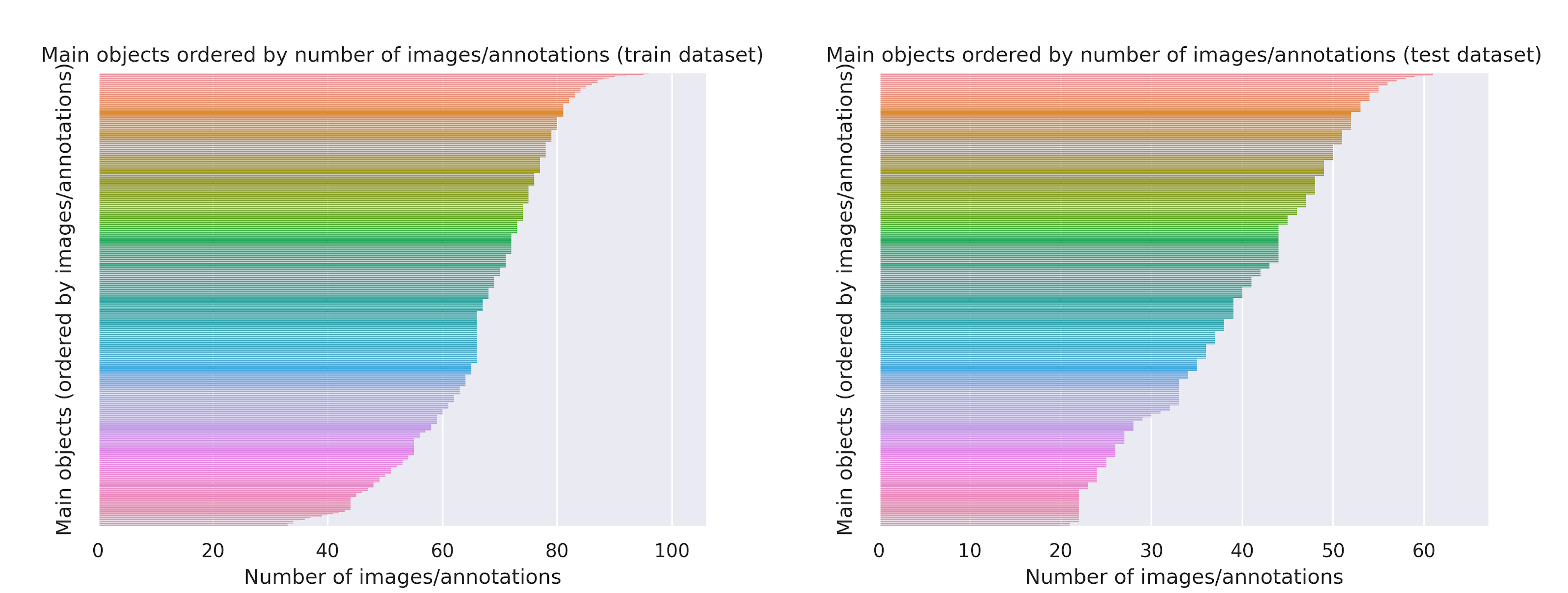}}
    \subfigure[Bar plots showing the number of main objects per category. The category with the most main objects is ``lamp''. For this plot, the train set and test set has the same distribution because the train/test split is obtained by allocating videos based on their main object.]{\includegraphics[width=0.95\columnwidth]{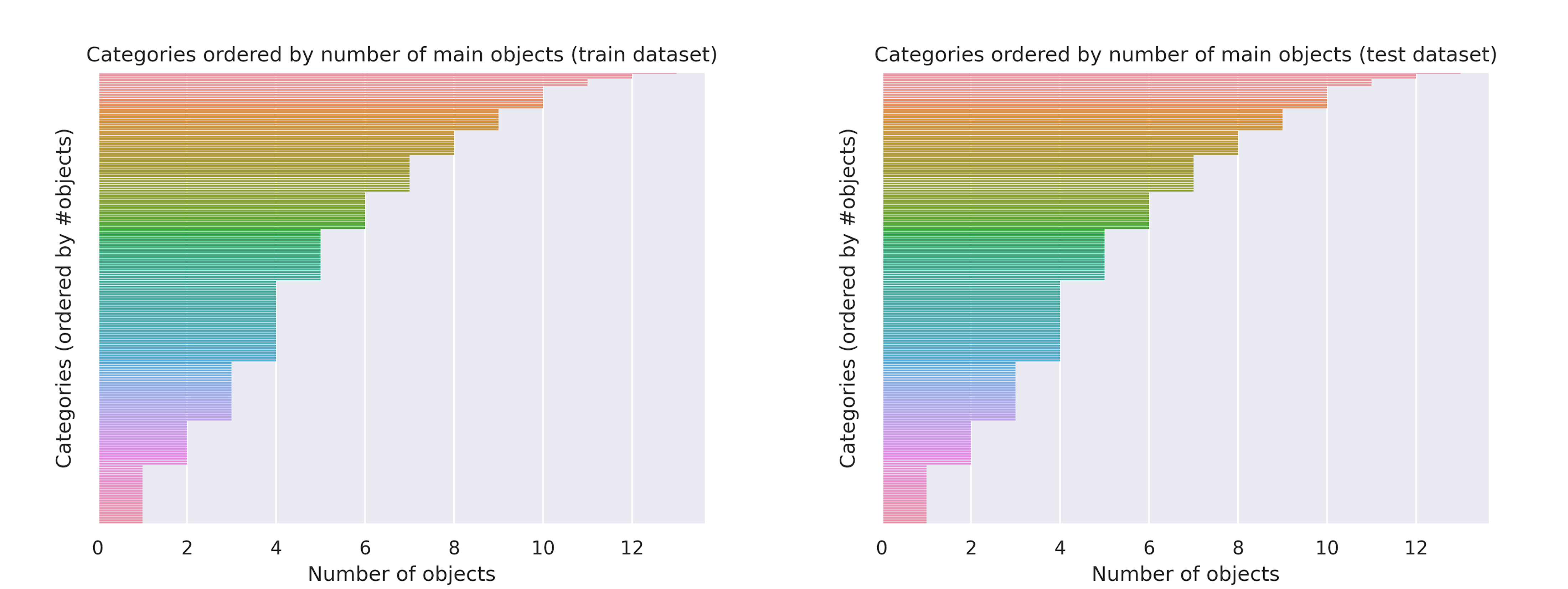}}
    \subfigure[Bar plots showing the number of objects per category in log scale. This includes all objects, even non-main ones. The category with the most objects is ``bottle''.]{\includegraphics[width=0.95\columnwidth]{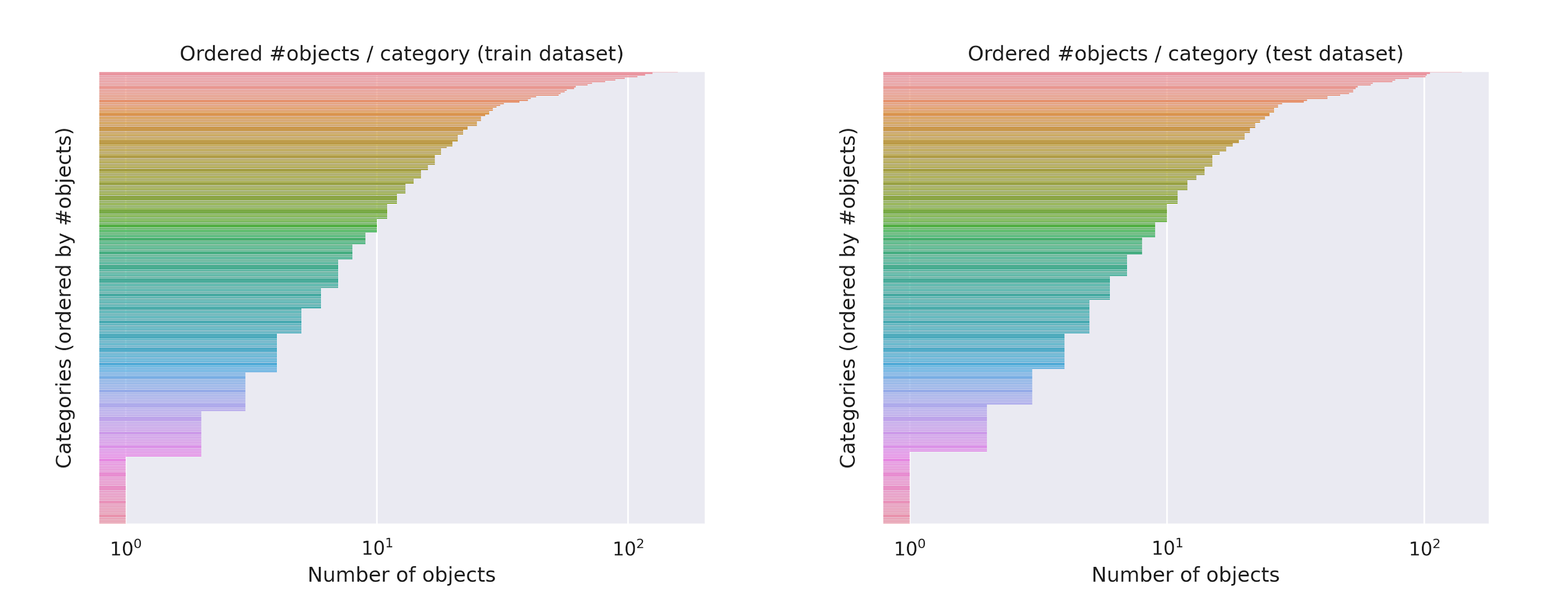}}
    \caption{Statistics of the EgoObjects challenge subset.}
    \label{fig:stats}
\end{figure}

The subset of EgoObjects used in the challenge contains around 100k video frames with $\sim$250k annotations, which correspond to 277 categories and 1110 main objects. There are 3407 and 3114 overall objects in the training set and testing set respectively. Additionally, the dataset follows a long-tailed distribution which makes the task more challenging and real-world oriented. We visualize some of the dataset's statistics in Figure \ref{fig:stats}. 

Compared to other egocentric datasets, the EgoObjects is both large-scale and object-centric. The most relevant to our dataset are those containing daily life objects, such as Objectron \cite{objectron}, CO3D \cite{co3d}, Epic-Kitchens \cite{epic-kitchens}, and Ego4D \cite{ego4d}. However, the Objectron and CO3D datasets have limited object categories, i.e. 9 and 50 respectively. Although Epic-Kitchens has 300 object categories, they only fall into the domain of kitchen activities. Additionally, all these datasets do not annotate the instance IDs. Ego4D, on the other hand, is the most large-scale egocentric dataset, but it is not object-centric.


\section{Competition}
\label{sec:competition}

The competition was organized using the two-phase structure following previous editions of the CLVision workshop~\cite{1st_clvision_challenge}: in the pre-selection phase, which took place between the 30th of March 2022 and the 29th of May 2022, participants were asked to submit their solution as a copy of the prediction obtained on the unlabeled test set; the second phase consisted in the finalist evaluation in which the top-ranking teams were asked to send their working code for a remote evaluation. The remote evaluation ensured a fair access to computational resources for all the participants.

\subsection{Challenge tracks}
\label{sec:tracks}
All tracks follow the common assumptions found in the continual learning literature: the continual learning system encounters a batch of new training data -- that in literature is usually named experience or task -- and the system can train only on that data without constraints on the number of epochs and strategy used. No access to previous or future training experiences is allowed (apart from the use of a limited replay memory). The system is evaluated at the end of each training experience on a separate test set. All tracks feature a task-agnostic (sometimes called Single Incremental Task) setup, which means that no task labels are provided both at train and test time.

The competition features three tracks all based on the EgoObjects dataset. The train and test sets are disjoint in the sense that (all frames of) a video can be found either in the train or test set, but not both. 

These three tracks are presented by their degree of difficulty and complexity.

\begin{enumerate}
    \item \textbf{Instance Classification.} In this track, images of the EgoObjects videos are cropped around the main object so that each image depicts a single object. For the class label, considering that EgoObjects features instance-specific labels, we use the instance labels instead of categories. This makes it an 1110 class classification problem in which classes belonging to the same semantic category share many visual clues. The continual learning benchmark is comprised of 15 class-incremental experiences. The number of classes is the same across all experiences.
    \item \textbf{Category Detection.} In this track, images from the dataset are used in the following way: multiple objects are depicted in common household and workplace environments and the goal is to predict the bounding box and category label of objects of known categories in each image. In each video, all annotations of all available classes (at the category level) are present, which makes it a data-incremental~\cite{delange2020a} benchmark. To provide structure to the benchmark, videos are arranged in 5 incremental experiences by applying the class-incremental ordering using the main object label. This however is one of the possible ways a data-incremental benchmark for continual object detection could be obtained. In particular, recent literature on continual object detection mainly focuses on annotation incrementality, in which the same images are encountered again in successive experiences but with a different set of annotations (usually class-incremental). We believe that this track provides a different, possibly more realistic angle on how benchmarks in this field can be built.
    \item \textbf{Instance Detection.} As with the Category Detection track, full images are given. However, each image carries the annotation of the main object only and the class label is the instance one. This hold for both train and test images. On one hand, one can be tempted to consider this track a simplification of the previous one, as the goal is to detect a single object in each image. However, the target object appears in scenes where objects from the same category are present making this easy for a detector to be tricked into detecting the wrong object. Moreover, considering that no information regarding the category of the main object is given (both at train and test time), an object of a completely different category could be mistakenly detected as the main object due to its similarity to a previously learned class. The benchmark features 5 class-incremental experiences. With this track, we tried to merge the complexity of instance-based recognition and object detection. For this, we deem this track as the most representative one of this competition.
\end{enumerate}

Participants were free to participate in all tracks in parallel. For this competition, tracks were evaluated separately, which means that the result obtained on a track did not impact the evaluation of other tracks. In addition, this also means that participants did not have to propose a unique algorithmic solution to be applied to all tracks: as tracks feature fairly different levels of complexity, we opted for simplicity and did not force participants to find a common solution. 


Monetary prizes were provided by Meta AI to stimulate participation. Prizes were split as follows (in USD):

\begin{itemize}
    \item \textbf{Instance Classification Track.} Winner: \$1500; 2nd Place \$900; 3rd Place: \$600.
    \item \textbf{Category Detection Track.} Winner: \$1700; 2nd Place \$1100; 3rd Place: \$700.
    \item \textbf{Instance Detection Track.} Winner: \$1700; 2nd Place \$1100; 3rd Place: \$700.
\end{itemize}

\subsection{Participation}
\label{sec:participation}

\begin{figure}[ht]
    \centering
    \includegraphics[width=0.8\columnwidth]{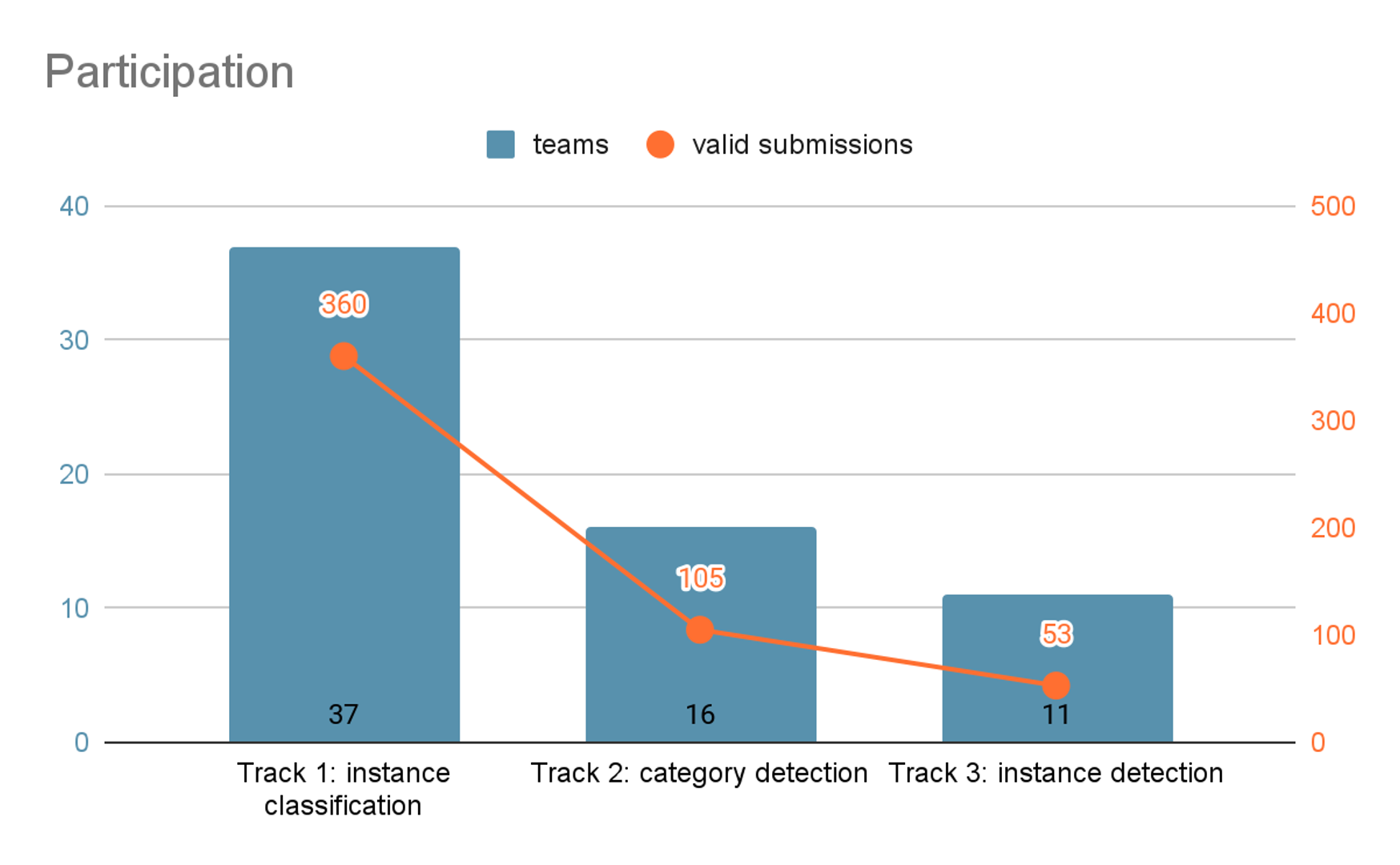}
    \caption{Visualization of teams and valid submissions across three tracks.}
    \label{fig:track_participation}
\end{figure}

The pre-selection phase was managed using the CodaLab public instance. Submissions for all tracks were opened simultaneously on the 30th of March and closed on the 29th of May. 
Figures regarding participation can be found in Table \ref{tab:participation_figures}, while Figures \ref{fig:ic_submissions}, \ref{fig:cd_submissions}, and \ref{fig:id_submissions} show the number of daily submissions for each track.

\begin{table}
\centering
\caption{Participation for the three competition tracks. Submissions were considered valid if uploaded results were aligned with the format expected by the scoring algorithm. The total number of registered and participating teams in the last row keeps into account that the same team registered and participated in multiple tracks.}
\label{tab:participation_figures}
\begin{tabular}{cllll} 
\hline
\textbf{\texttt{\#}} & \textbf{Track Name}     & \begin{tabular}[c]{@{}l@{}}\textbf{Registered}\\\textbf{Teams}\end{tabular} & \begin{tabular}[c]{@{}l@{}}\textbf{Participating }\\\textbf{Teams}\end{tabular} & \begin{tabular}[c]{@{}l@{}}\textbf{Valid}\\\textbf{Submissions}\end{tabular}  \\ 
\hline
1           & Instance Classification & 60                                                                          & 37                                                                              & 360                                                                           \\
2           & Category Detection      & 46                                                                          & 16                                                                              & 105                                                                           \\
3           & Instance Detection      & 34                                                                          & 11                                                                              & 53                                                                            \\ 
\hline
            &   Total (unique)                      & 77                 & 47                                                                             & 518                                                                          
\end{tabular}
\end{table}

The competition attained good participation levels, especially on the Instance Classification track. For the harder “Instance Detection” track, most submissions were received near the end of the pre-selection phase. A more in-depth analysis revealed that most submissions came from teams in the higher part of the leader board of the similar “Category Detection” track. An in-depth description of the tracks is given in Section \ref{sec:tracks}. Considering the limited literature in the continual object detection field, detection tracks saw a surprising overall high participation both in terms of registered teams and the number of valid submissions with 158 overall submissions across the detection tracks and 20 unique participating teams.


\subsection{Metrics}
\label{sec:metrics}
Scores are computed using the \textit{Average Mean Class Accuracy} (AMCA) metric on all tracks, which is a well-known metric being adopted in recent works in the CL field. The same metric was also used in the SSLAD competition \cite{sslad_challenge}.
Following the notation in \cite{amca_eq}, the AMCA metric is defined as:

\begin{gather}
\textnormal{AMCA} = \frac{1}{N_c \cdot N_{e}} \sum_{c=1}^{N_c} \sum_{e=1}^{N_{e}} \frac{correct_e^c}{total_e^c}
\end{gather}

where $N_c$ is the overall number of classes, $N_{e}$ is the number of training experiences, $correct_e^c$ and $total_e^c$ are the number of correctly classified test images of class $c$ on experience $e$, and the total number of test images of class $c$ in experience $e$, respectively. This is preferable to the plain final accuracy as it equally weights the scores obtained after each training experience. In addition, differently from the plain average of test accuracies, the accuracy obtained for different classes is first averaged, which is particularly important when an imbalanced test set is used. The same idea can be applied to detection tasks by averaging the Mean Average Precision (mAP) scores across experiences\footnote{The Average Precision is defined following the COCO convention (AP at IoU=.50:.05:.95).}. We name this metric \textit{Experience Average Precision} (EAP).

\subsection{Rules and Evaluation}
\label{sec:rules_and_eval}

The goal of a challenge is to encourage participants to explore novel solutions to the proposed problem. Rules (and in some way metrics) restrict the kind of acceptable solutions and thus limit the degree of novelty the challenge aims to achieve. The idea of “imposing a direction” to solutions may be a good idea when the goal is to stimulate contributions more focused on optimizing specific aspects (i.e., compute performance, memory efficiency, training latency). 
However, this challenge was organized with the idea of stimulating innovation by allowing for maximum freedom. 
Because of this, only a minimal set of rules were enforced. These rules are needed to define a fair common ground for all participants, but they do not impose constraints on the continual learning strategies to employ.

\begin{itemize}
\item \textbf{Maximum replay buffer size}. A maximum replay buffer size is needed to prevent overuse of past data. The choice of the maximum size was based on track complexity: for tracks 1 and 2 it was set to 3500 exemplars, while for track 3 it was set to 5000.
\item \textbf{Pretraining}. Pretraining was allowed on the mainstream datasets ILSVRC-2012, COCO2017, and LVIS. No additional pre-training data was allowed, and the replay buffer cannot contain data from pretraining datasets.
\item \textbf{Model size}. A maximum model size of 70M parameters was enforced. A constraint regarding the maximum model size is needed to i) compare the continual learning solution on a fair basis (especially considering that pre-training is allowed); ii) prevent participants with more compute power to obtain an unfair advantage from using bigger models. It must be noted that the goal of a continual learning challenge is to propose a winning continual learning strategy: given a reasonable maximum model size, the continual learning strategy should be able to incrementally learn using a model of that size (or less). At the same time, no constraints on the model architecture were enforced to allow for architectural solutions to be adopted.
\item \textbf{No test time training}. Another decision regards the ability to tune the model at test time. Test time training (or tuning) refers to an approach in which the model is tuned to boost its performance on test examples \cite{pmlr-v119-sun20b}. While test time training is an interesting research direction, for this challenge we wanted to model a more classic setup in which the incrementally trained system should be immediately usable for inference, with minimal latency and compute cost, after the incremental training phase.
\item \textbf{Time limit}. In addition, a time constraint was enforced to prevent extreme solutions from being proposed. The time limit for the classification track has been set to 12 hours, while for detection tracks the time limit was set to 24 hours. This time refers to the complete execution of training and test phases for all the incremental experiences. Similar to the rule regarding the model size, enforcing a time limit has a similar positive effect on fairness by reducing the advantage of participants who are able to use massive computing power for their experiments.
\end{itemize}

A starter development kit (devkit) was made publicly available as a git repository \cite{challenge_repo}. The devkit was based on the Avalanche \cite{Lomonaco_2021_CVPR} continual learning library. The devkit already packaged an Avalanche-based template that could be used as a starting point to run alternate training and evaluation phases in the correct order. The only constraint regarding the devkit was the data loading component, which the participants could not change. All the other parts could be customized at will, including the choice of frameworks and libraries to employ.

\subsection{Solutions}
In this section we provide a summary of the solutions from the winners of each track. These approach summaries are based on the analysis of the report and the source code provided by each team. Full-length reports can be found on the official repository of the competition \cite{challenge_repo}. Intermediate scores for the winning solutions are given in Tables \ref{tab:winners_scores_track1}, \ref{tab:winners_scores_track2}, and \ref{tab:winners_scores_track3}.

\subsubsection{Track 1: Instance Classification}
As described in Section \ref{sec:tracks}, this is the only classification track in the competition. The provided solutions should be able to learn to distinguish 1110 unique objects depicted in different poses and distances under varied lightning conditions and background scenarios. The training stream is composed of 15 incremental experiences and no task labels are provided both at train and test time.

\begin{figure}[ht]
    \centering
    \includegraphics[width=0.8\columnwidth]{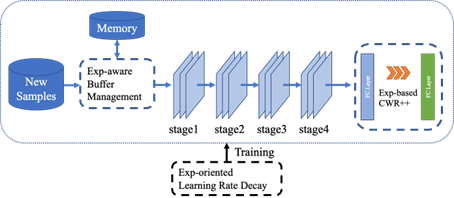}
    \caption{The overall pipeline of the solution from Tencent Youtu Lab.}
    \label{fig:track1_youtu1}
\end{figure}

\noindent\textbf{Tencent Youtu Lab (Rank 1)}
The overall pipeline of the solution is shown in Figure \ref{fig:track1_youtu1}. Firstly, the training data of the current experience is generated by an Experience-aware Buffer Management module, that keeps more samples of older experiences. After that, the training of the feature extractor and classifier is performed. The weight fusion between experiences is performed by Experience-based CWR++ (a novel extension of CWR+ \cite{MALTONI201956}) after the training is finished. The training process uses an Experience-oriented Learning Rate Decay to alleviate the catastrophic forgetting among experiences.

The solution exploits the Twins-SVT-Base \cite{twins} as backbone model
\footnote{The pretrained weights were obtained from Timm: \url{https://github.com/rwightman/pytorch-image-models}}. 
In addition to the CE loss, the network was trained using label smoothing \cite{Szegedy_2016_CVPR}.
They use Albumentations \cite{albumentations} to perform most data augmentations, including shift scale rotate, blur, distortion, and noise. Due to the fact that it was difficult to distinguish whether some cases were the same instance or not, they let the model exploit brightness and contrast hue to assist in classification by turning off some augmentations.

\noindent\textbf{BaselineCL (Rank 2)}
Pre-trained models on large-scale datasets can quickly and well transfer to other datasets. Considering that the competition dataset is relatively small, it is easy to overfit when learning from scratch.
Therefore, they focused on evaluating a series of competitive pre-trained classification models. For their evaluation, the BaselineCL team considered ResNet \cite{He_2016_CVPR}, 
EfficientNet \cite{pmlr-v97-tan19a}, 
and RegNet \cite{Radosavovic_2020_CVPR}. 

\begin{figure}[t]
    \centering
    \includegraphics[width=0.7\columnwidth]{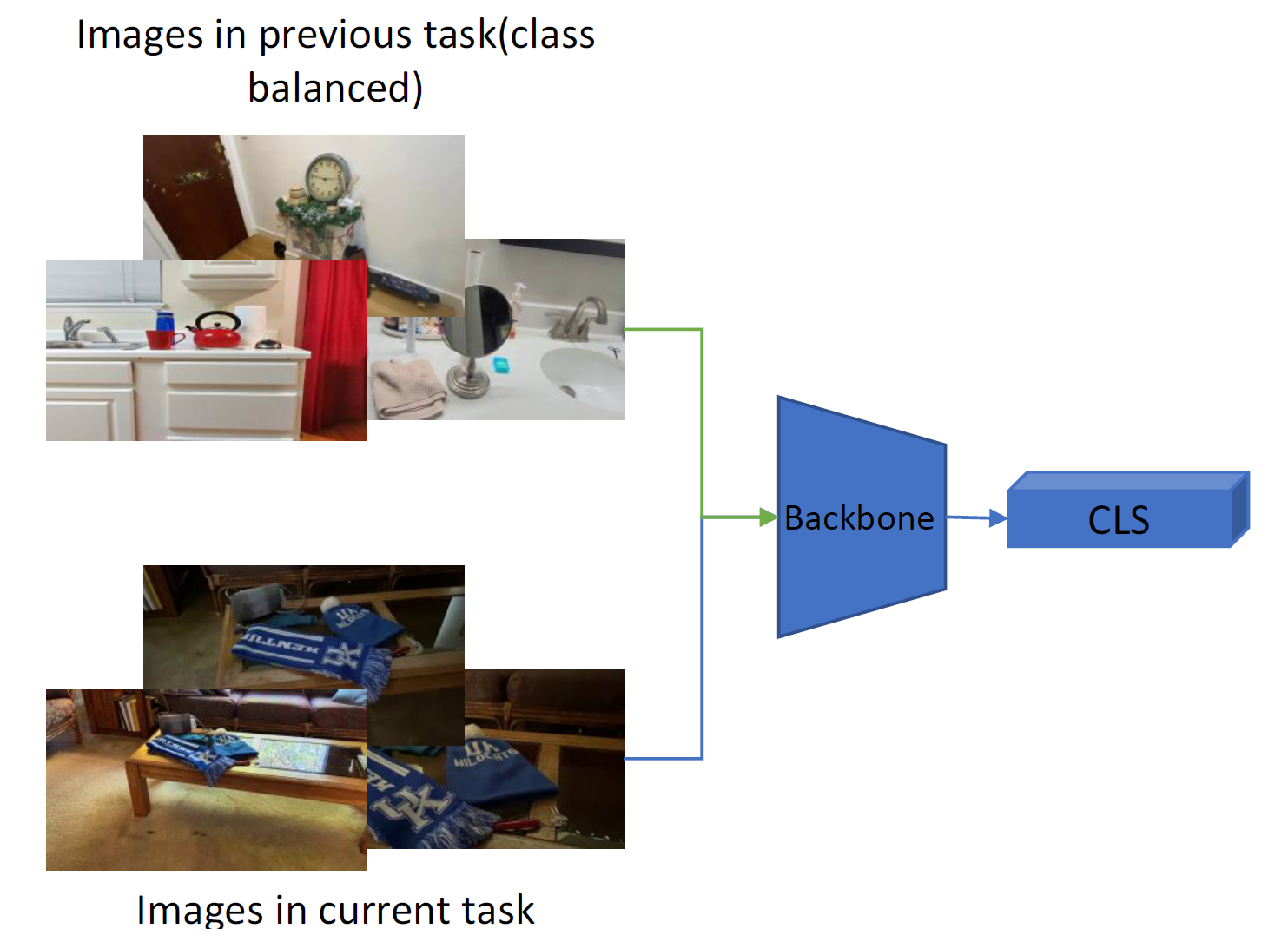}
    \caption{BaselineCL: the classes-balanced replay strategy.}
    \label{fig:track1_baseline1}
\end{figure}

For the incremental learning component, 
they focus on reducing the forgetting of the old categories by using a class-balanced replay strategy (shown in the Figure \ref{fig:track1_baseline1}), which trains the network by retaining some of the learned knowledge and combining it with the training data of the current task.
Considering the category imbalance in the learned data, they select the learned samples according to the category to attenuate the effect of category imbalance on the training results.

At each incremental experience, a RegNet \cite{Radosavovic_2020_CVPR} is trained using SGD with 3 epochs. The learning rate is adjusted according to a cosine decaying policy.



To accelerate the rate of convergence during model training and make the training process more stable, they introduce a Batch Normalization (BN) layer before the fully-connected layer. They found that backbones with such BN layer outperform others. 


To avoid overfitting and improve the generalization ability of model, they use label smoothing CE Loss (LSCE) \cite{Szegedy_2016_CVPR} to train the model, which they find to be beneficial in terms of final accuracy. 


To choose the continual learning strategy to use, they evaluated different existing techniques (and their combination) such as LwF \cite{Li2016}, EWC \cite{Kirkpatrick2016}, SI \cite{Zenke2017}, GDumb \cite{prabhu2020}, MAS \cite{Aljundi_2018_ECCV}, iCaRL \cite{rebuffi2017icarl}, and Replay (with and without class balance). They find that that a class balanced replay strategy is very efficient even when used alone. To carry out these experiments, they use the training set as the test set (as the labels for the challenge test set are not available).


\noindent\textbf{SolangKim (Rank 3)}
For their solution, the SolangKim team focused on preventing forgetting by applying several continual learning algorithms such as A-GEM \cite{agem}, EWC \cite{Kirkpatrick2016}, Synaptic Intelligence \cite{Zenke2017}, and replay.

For the replay strategy, they used a strategy that keeps the ratio of each experience equal, and the memory size of the replay buffer used in each experience is fixed to 2000 samples. When it comes to regularization, a mix of EWC, SI, and AGEM was used, by taking their implementations from Avalanche \cite{Lomonaco_2021_CVPR}.

For the model, they used the timm \cite{rw2019timm} library to check the performance of various backbones. According to \cite{Mirzadeh_architecture}, in continual learning settings, increasing the width is more effective in preventing catastrophic forgetting than increasing the depth of the network. Therefore, they used a WideResNet50\_2 (ResNet50 with 2 width multiplier) implemented in timm. This showed better performance compared to ResNet50 and ResNet101. The encoder was initialized using ImageNet pre-trained weights. The classifier was left randomly initialized. For the encoder part, a small learning rate ($0.001$) was used, as pre-trained weights are employed. In the classifier part, a higher learning rate ($0.01$) was used because of randomly initialized weights. A cosine learning rate scheduler was used with the SGD optimizer. Batch normalization was used only in the first experience, and batch statistics were fixed from the next experiences. This is because the newly learned statistics are different from the statistics of previous experiences, which can cause forgetting.

\begin{table}[H]
\scriptsize
\setlength\tabcolsep{2pt}
\centering
\caption{Instance Classification (Track 1): mean test accuracy after each experience and final AMCA.}
\label{tab:winners_scores_track1}
\begin{tabular}{lllllllllllllllll} 
\hline
Team & $E_0$ & $E_1$ & $E_2$ & $E_3$ & $E_4$ & $E_5$ & $E_6$ & $E_7$ & $E_8$ & $E_9$ & $E_{10}$ & $E_{11}$ & $E_{12}$ & $E_{13}$ & $E_{14}$ & $AMCA$ \\ 
\hline
Tencent Youtu Lab & 0.0666 & 0.1312 & 0.1948 & 0.2616 & 0.3271 & 0.3918 & 0.4584 & 0.5225 & 0.5895 & 0.6558 & 0.7203 & 0.7842 & 0.8477 & 0.9094 & 0.9740 & 0.5223 \\
BaselineCL & 0.0663 & 0.1316 & 0.1964 & 0.2623 & 0.3283 & 0.3934 & 0.4593 & 0.5232 & 0.5890 & 0.6542 & 0.7156 & 0.7788 & 0.8427 & 0.9016 & 0.9655 & 0.5205 \\
SolangKim & 0.0593 & 0.1307 & 0.1923 & 0.2590 & 0.3252 & 0.3891 & 0.4542 & 0.5183 & 0.5850 & 0.6507 & 0.7163 & 0.7791 & 0.8416 & 0.9043 & 0.9690 & 0.5183 \\
\hline
\end{tabular}
\end{table}
\begin{table}[H]
\centering
\caption{Category Detection (Track 2): mAP after each experience and final EAP.}
\label{tab:winners_scores_track2}
\begin{tabular}{lllllll} 
\hline
Team & $E_0$ & $E_1$ & $E_2$ & $E_3$ & $E_4$ & $EAP$ \\ 
\hline
Youtu-Fuxi-det & 0.3061 & 0.4723 & 0.5814 & 0.6753 & 0.7618 & 0.5594 \\
NUS-LVLAB\&ASTAR & 0.2836 & 0.4470 & 0.5762 & 0.6789 & 0.7818 & 0.5535 \\
USTC & 0.1951 & 0.3445 & 0.4391 & 0.5265 & 0.6151 & 0.4241 \\
\hline
\end{tabular}
\end{table}
\begin{table}[H]
\centering
\caption{Instance Detection (Track 3): mAP after each experience and final EAP.}
\label{tab:winners_scores_track3}
\begin{tabular}{lllllll} 
\hline
Team & $E_0$ & $E_1$ & $E_2$ & $E_3$ & $E_4$ & $EAP$ \\ 
\hline
Youtu-Fuxi-det & 0.2330 & 0.3953 & 0.5459 & 0.7021 & 0.8560 & 0.5465 \\
NUS-LVLAB\&ASTAR & 0.1505 & 0.3044 & 0.4550 & 0.6079 & 0.7537 & 0.4543 \\
Angelo Menezes & 0.1474 & 0.2909 & 0.4232 & 0.5539 & 0.6691 & 0.4169 \\
\hline
\end{tabular}
\end{table}

\
\noindent\textbf{Track 1: Discussion}
For the first track, each participant mainly focused the investigation on a different element of the overall solution leading to different reasonings. The winning team mostly focused on designing a mechanism to properly stabilize training on the last layer (CWR++), augmentations, and input preprocessing. This solution is arguably the most novel in terms of algorithmic contributions.
The runner-up team performed an empirical analysis of the model architecture to use among many alternatives. Then, using the model selected, they tried mixing many existing continual learning techniques. The final result is that a properly chosen model and a replay strategy can obtain good results.
Finally, the third-place team focused more on choosing the correct mix of techniques which resulted in a complex hybrid strategy that include the use of concurrent regularization techniques. The choice of the model was based on previous studies \cite{Mirzadeh_architecture} that showed how wide CNN models outperform deep ones in continual learning tasks. 

However, there are elements that are recurrent in these solutions. Firstly, all participants employ a replay buffer of the maximum allowed size. This is an expected outcome as this choice has no impact on the final score. However, using replay means that time must be spent to train on replay data, which may be a problem considering the time limit imposed by the rules. The same time could be used to apply more complex algorithms. These solutions, as well as the empirical findings of previous competitions, confirm that a properly tuned replay strategy is arguably the most important element to prevent forgetting and thus is worth the memory and time required to apply it. All the solutions use some form of buffer balancing policy, with different policies possibly leading to different outcomes. In particular, the winning team found out that keeping more samples of older experiences is beneficial to contrast forgetting. This result may not hold in setups different from the class-incremental one.

Secondly, both the second and third place participants adopted specific strategies to handle the training of Batch Normalization (BN) layers. While BaselineCL (2nd place) found out that introducing a BN layer before the last layer is beneficial, SolangKim (3rd place) preferred to freeze batch normalization statistics after the first experience.
Both Tencent Youtu Lab (1st place) and BaselineCL employ label smoothing, which they found to have a beneficial effect on retaining past knowledge. In addition, both teams use replay along with techniques that do not apply regularization to the entire model.

Surprisingly, the top-3 teams did not use distillation techniques. Distillation techniques are widely explored in the continual learning literature and have been successfully used by top-performing teams of detection tracks.

When it comes to the architecture to adopt, the winner employed a vision transformer. However, the results obtained from the other teams show that classic CNN architectures can be used with success on this task.

\subsubsection{Track 2: Category Detection}

As mentioned earlier, the goal of this track is to predict the bounding box and category label of objects of known categories in each image, with training and testing data presented with 5 incremental experiences in a class-incremental order.  

\begin{figure}
    \centering
    \includegraphics[width=0.9\columnwidth]{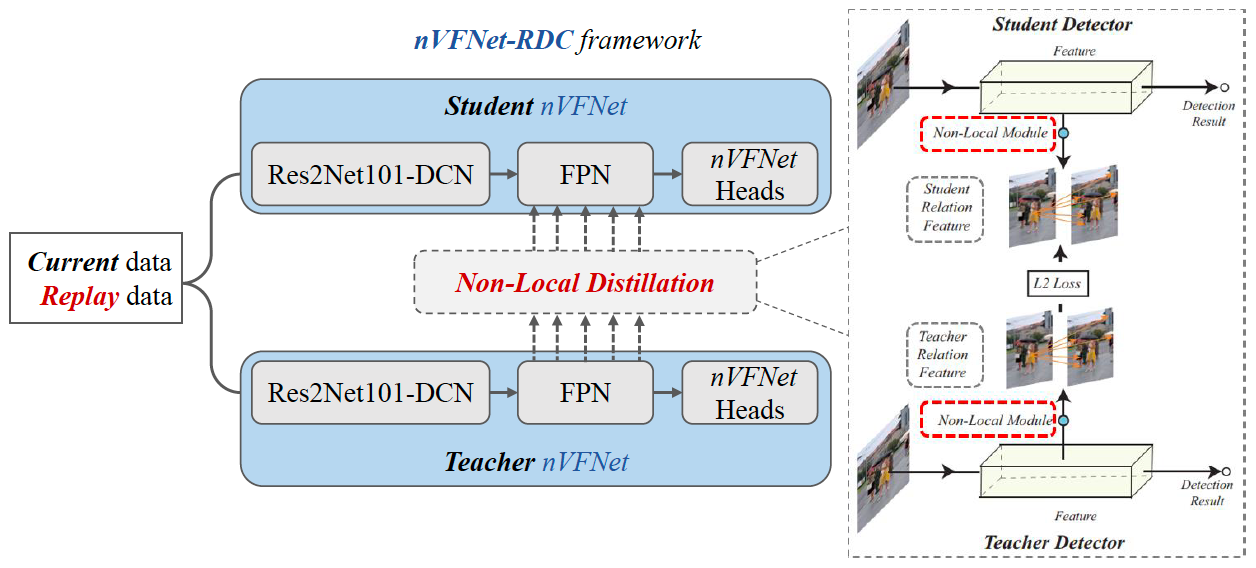}
    \caption{The framework of the nVFNet-RDC from Tencent Youtu Lab. Upon the proposed nVFNet detector, the continual learning approaches containing Replay and Non-Local Distillation are integrated.}
    \label{fig:nvfnet_rdc}
\end{figure}

\begin{figure}
    \centering
    \includegraphics[width=0.9\columnwidth]{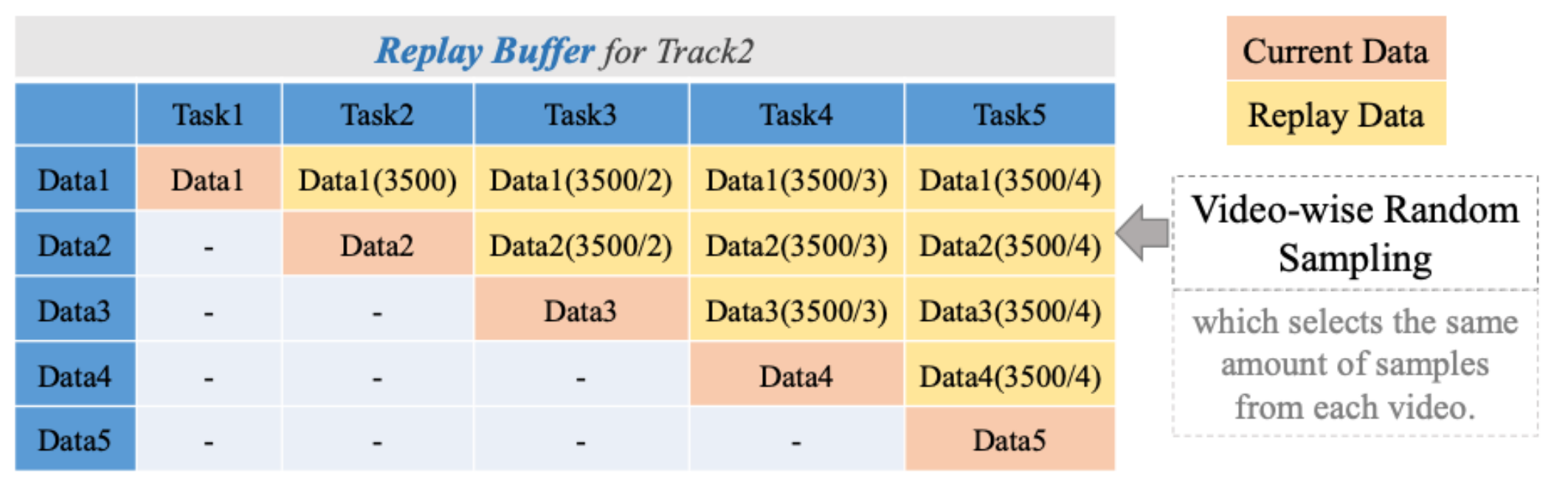}
    \caption{
        \textbf{Replay strategy for Tencent Youtu Lab}: Task1-5 and Data1-5 denote 5 incremental tasks and their corresponding training sets. 
        The training data for each task is shown in each column, which contains current data (orange) and replay data (yellow). 
        The same amount of data is sampled from previous tasks using video-wise random sampling. 
    }
    \label{fig:nvfnet_replay}
\end{figure}

\noindent\textbf{Tencent Youtu Lab (Rank 1)}
This submission proposes a framework dubbed as nVFNet-RDC, whose overall architecture is illustrated in Figure~\ref{fig:nvfnet_rdc}. 
The framework features a teacher-student distillation setup where the detector trained from previous experiences is used as the teacher to distill information to the student detector that's under training in current experience. Specifically, the distillation process happens between the outputs of two non-local module, in order to capture the relationship between instances. As the base detector, the proposed framework adopts the one-stage VFNet~\cite{varifocalnet}. 

To cache samples from previous experiences, the proposed method adopts the replay rules illustrated in Figure~\ref{fig:nvfnet_replay}. 
Task1 to Task5 are 5 incremental tasks. Data1 to Data5 denote training sets for the corresponding tasks. 
The training data for each task is shown in each column, which contains current data (orange) and replay data (yellow). 
The same amount of data is sampled from previous tasks using video-wise random sampling. 

\noindent\textbf{NUS \& A*STAR (Rank 2)}
As shown in Figure~\ref{fig:xingyi_frcnn}, the proposed architecture features a shared backbone, and 5 heads that are separately trained for 5 different experiences. Each of these 5 heads contain a set of RPN and ROI heads. The proposed architecture uses ResNet50 as the backbone and adopts standard RPN and ROI head design with some modifications.    
Specifically, to reduce the model size, the authors replaced the fully connected layer that follows the ROI pooling with a simple average pooling. Meanwhile, they substituted some vanilla convolutions with deformable convolutions~\cite{dai-iccv2017}, and standard RPN with Guided Anchoring RPN~\cite{wang-cvpr2019}, to further boost the model accuracy. 

The proposed method adopts a ``Minority Replay'' strategy. 
Specifically, at the beginning of each experience, they count the instance number for each category in the training sample and previous replay buffer. 
Then, they sort the class ID by their instance counts and pick the top 3500 images from the minority class and place them into the pool for the next iteration.

\noindent\textbf{USTC (Rank 3)}
This submission uses a Faster-RCNN detector with a Swin-Small transformer backbone, as they have compared across several backbone architectures (ResNet50, Swin-Tiny, Swin-Small) and found the best tradeoff with Swin-Small. 
As for the replay strategy, in each training experience, a certain number of samples are randomly selected in the experience, and shuffled into the pool such that they maintain a constant size of 5000 samples for the replay buffer. 

\noindent\textbf{Discussion}
We can summarize the participating approaches from the following aspects. 
First, when dealing with the catastrophic forgetting, both the \textit{Tencent Youtu Lab} and \textit{USTC} teams adopted a uniform sampling strategy to cache data from previous tasks, with the difference in that the former approach samples history data in a video-wise manner while the latter did so on a frame basis. Unlike these two teams, \textit{NUS \& A*STAR} team proposes to follow a ``Minority Replay'' strategy in which they sort and emphasis on selecting and buffering data from rare classes.  
In addition to replay buffer, the participants added further designs to mitigate catastrophic forgetting --- while \textit{Tencent Youtu Lab}  proposes to distill information from a teacher model trained in previous tasks to the model for the current task, \textit{NUS \& A*STAR} proposes to train separate detection heads for different tasks. 
Most teams simply borrow established detectors from the standard detection setting. For example, \textit{Tencent Youtu Lab} adopts the VFNet~\cite{varifocalnet} one-stage detector, whereas the other two teams adopt the well-tested two-stage Faster-RCNN style detectors. 
As for the backbone network, most teams adopted the classical ResNet CNNs with the exception from \textit{USTC}, where they ablated the comparisons between ResNets and Transformer architectures and chose Swin-Small as their backbone network. 

In a nutshell, despite many insightful designs and detailed ablations, the winning methods for category detection in the continual learning setting still mostly lies in the boundary of existing techniques tailored for the standard non-continual setting. This means that there are still huge potential for innovative ideas in this field, and it would be exciting to witness how this rapidly evovling field moves in the next few years.   

\subsubsection{Track 3: Instance Detection}

\noindent\textbf{Tencent Youtu Lab (Rank 1)} The proposed method is called nVFNet-RDC. As shown in Figure \ref{fig:nvfnet_rdc}, the nVFNet-RDC framework consists of teacher-student models, and adopts replay and feature distillation strategies. The proposed nVFNet Detector combines VFNet \cite{varifocalnet} and Non-Local Dense Classifier. Additionally, a continual learning approach named RDC is proposed, which consists of Replay and Non-Local Distillation. Other Components include: Res2Net \cite{res2net} backbone pre-trained on COCO \cite{coco_dataset}, and PhotoMetricDistortion data augmentation.

With the replay strategy, the input data contains current data and replay data. There is a teacher model trained from the previous task, which consists of Res2Net backbone, FPN \cite{fpn} neck and detector heads. And the student model needed to be trained for current task, which has the same structure as the teacher model. Along with the conventional supervised loss, a Non-Local Distillation is applied to the features generated by FPN.

\noindent\textbf{NUS \& A*STAR (Rank 2)}
This team proposed a multi-head Faster-RCNN \cite{faster-rcnn} with heuristic data reply strategy, with the goals of 1) achieving strong performance without forgetting the previous experiences; 2) performing robustly to data quality and variation, and 3) not exceeding the required model size.

\begin{figure}
    \centering
    \includegraphics[width=0.8\columnwidth]{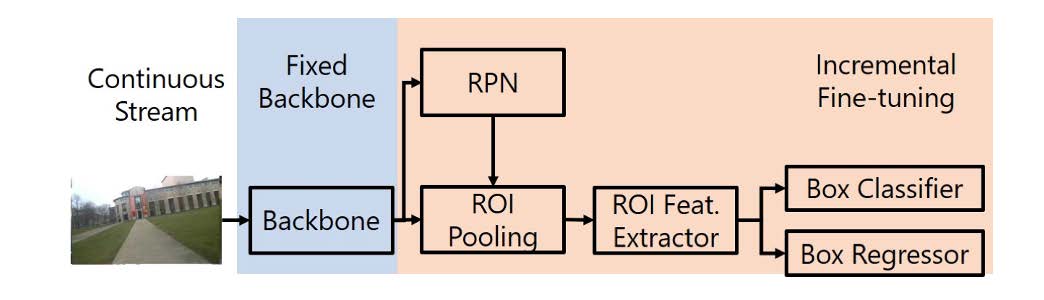}
    \caption{Faster RCNN with fixed and shared backbone and multiple detection heads incrementally fine-tuned per experience.}
    \label{fig:xingyi_frcnn}
\end{figure}

As shown in Figure \ref{fig:xingyi_frcnn}, a new detector head is used to deal with the new data at each experience. Using distinct modules to store the knowledge of prior experiences, largely alleviates the catastrophic forgetting problem. The unshared part includes the FPN neck, the RPN head, and the ROI head. Only the ResNet-50 backbone is shared across experiences and kept frozen during training. The model has $T$ detector heads where $T$ is the number of experiences. At the training time for experience $t$, only the $t$-th head is updated, while other parameters are untouched. At the $t$-th evaluation, the predicted bounding boxes of the 1, . . . , $t$-th head are merged by NMS. For the replay strategy, they randomly sample one frame from each video sequence and fill the full replay buffer.

\noindent\textbf{Angelo Menezes (Rank 3)} For the architecture, FCOS \cite{fcos} is chosen with ResNet-50 \cite{He_2016_CVPR} and FPN \cite{fpn}. The key design includes two components. One is the balanced experience replay, where each instance has same number of samples for the previous experiences. The other one is knowledge distillation from features and outputs. Following the basic distillation procedure for incremental object detection introduced by Shmelkov et al. \cite{inc_det} and the following advances proposed by Chen et al. \cite{kd_incdet}, the learning of each new experience is regularized by distilling knowledge from a saved version of the model trained on the previous experiences. The distilled knowledge comes from the $L_2$ loss, applied to the detection head and intermediate features in the backbone.

\noindent\textbf{Discussion}
To summarize, the continual learning of the instance-level object detection is an under-explored area. The top submissions basically follow the similar strategy of applying continual learning techniques to a base category-level object detector and treating each instance as an individual category. For the base detector, both single-stage and two-stage detectors are applicable. Tecent Youtu Lab (1st place) and Angelo Menezes (3rd place) are using single-stage detectors (VarifocalNet \cite{varifocalnet} and FCOS \cite{fcos}) while NUS\&A*STAR (2nd place) is using two-stage detector (Faster R-CNN \cite{faster-rcnn}). And the detectors are pretrained on either COCO \cite{coco_dataset} or LVIS \cite{lvis_dataset} which are not the major factor of performance. For the network architecture, ResNet \cite{He_2016_CVPR} and its variants \cite{res2net} are still the dominating backbones. One key factor contributing to the performance is the replay buffer. All submissions observe that the use of full replay buffer is necessary for high performance. But they adopt different sampling strategies for filling the buffer. Specifically, Tecent Youtu Lab (1st place) samples an experience-balanced buffer, whereas NUS\&A*STAR (2nd place) and Angelo Menezes (3rd place) uses video-balanced buffer and instance-balanced buffer respectively. Distillation techniques are also widely used by 1st and 3rd places.
We summarize these design choices of top three submission in Table \ref{tab:track3_compare}.

\begin{table}
\setlength\tabcolsep{3pt}
\centering
\caption{
    Comparison between the top 3 submissions for Track 3.
}
\label{tab:track3_compare}
\begin{tabular}{l|lllll|ll}
\hline \hline
Team             & Base Detector & Backbone       & Pretrain & Repaly              & Distillation & AP   & AP50 \\ 
\hline
Tecent Youtu Lab (1st)      & VarifocalNet  & Res2Net101 & COCO     & Experience & ~~~$\checkmark$ & 54.7 & \hspace{0.4mm} 61.2 \\
NUS\&A*STAR (2nd)     & Faster R-CNN  & ResNet50       & LVIS     & Video      &              & 45.4 & \hspace{0.4mm} 56.0 \\
Angelo Menezes (3rd)      & FCOS          & ResNet50       & COCO     & Instance   & ~~~$\checkmark$ & 41.7 & \hspace{0.4mm} 51.1 \\ 
\hline
\end{tabular}
\end{table}

\section{Conclusions}
The 3rd Continual Learning for Computer Vision Competition has been an excellent opportunity to introduce the complex task of detecting objects from egocentric video streams. With the introduction of a purposefully built dataset covering 1110 different objects, the challenge tried to solicit participants in designing novel continual learning approaches that can operate in realistic object classification and detection continual learning tasks both at the category and at the instance level.

The competition saw significant participation on all tracks, with detections tracks having received 105 (Category Detection) and 53 (Instance Detection) valid submissions. The Instance Classification track remains the most popular one with 360 valid submissions.
The idea of offering classic classification tracks along with novelty tracks is now being more and more consolidated, with other recent competitions (such as the 2nd CLVision competition \cite{2nd_clvision_challenge} and the SSLAD competition \cite{sslad_challenge}) having successfully embraced this setup. We believe that this approach can further increase participation in novelty tracks.

Despite the good participation figures and the results achieved, room for improvement exists in the following areas:

\begin{itemize}
    \item \textit{Consider performance metrics.} Performance metrics, such as memory and computing time, have not been considered for this competition. However, considering performance metrics in the final score may solicit more diverse solutions. This is especially important if the goal is to move towards more realistic scenarios: continual learning solutions that can operate with limited working/storage memory and run in a minimal time may be preferable to accuracy-maximizer ones.
    \item \textit{Test on multiple variations of the benchmark.} Evaluating the solutions on variations of the same benchmark may prevent participants from overly designing solutions on the provided benchmark. To do this, the finalists selection phase could be done on a single variation of the benchmark, as it was done in the present competition. However, the finalists evaluation phase could be run on pre-defined variations of that benchmark. How variations can be introduced on a certain continual learning benchmark depends on the benchmark itself. For instance, in class-incremental setups variations can be crafted by changing the order of classes encountered in time. In addition to evaluating solutions on multiple benchmark variations, the idea of using an additional held-out test set should be considered.
\end{itemize}







\section*{Acknowledgements}
We would like to thank all the 3rd Continual Learning in Computer Vision workshop organizers, challenge chairs, and participants for making this competition possible. We also like to acknowledge workshop sponsors ContinualAI, MATRIX AI Consortium, ServiceNow and NVIDIA for their support in the organization of the workshop at CVPR 2022. We also would like to thank Meta AI for providing the dataset used in the challenge.

\appendix

\section{Participation over time}\label{sec:participation_over_time}

\begin{figure}[ht]
    \centering
    \includegraphics[width=0.80\columnwidth]{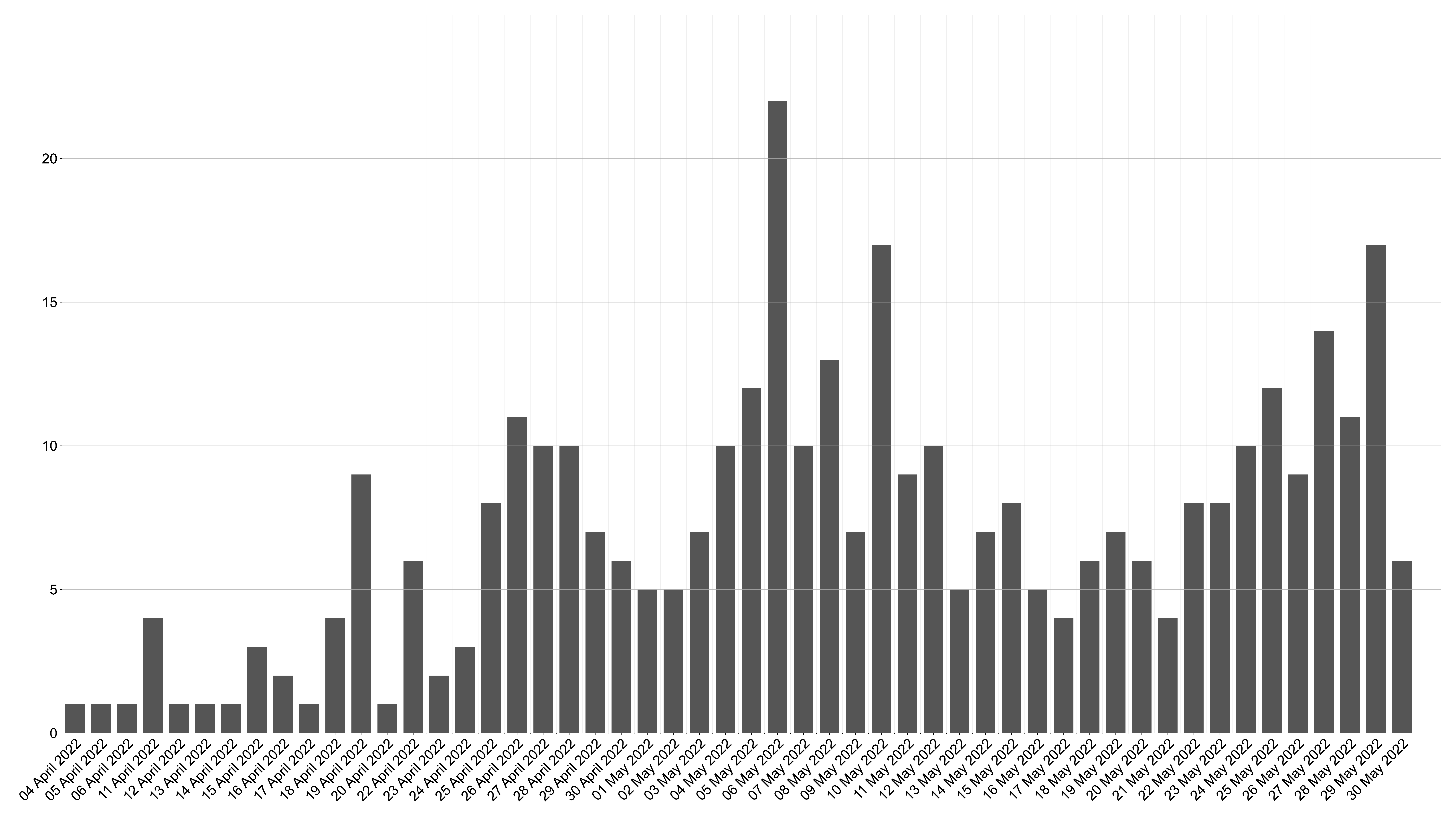}
    \caption{Daily submissions for the Instance Classification track (Track 1).}
    \label{fig:ic_submissions}
\end{figure}

\begin{figure}[ht]
    \centering
    \includegraphics[width=0.80\columnwidth]{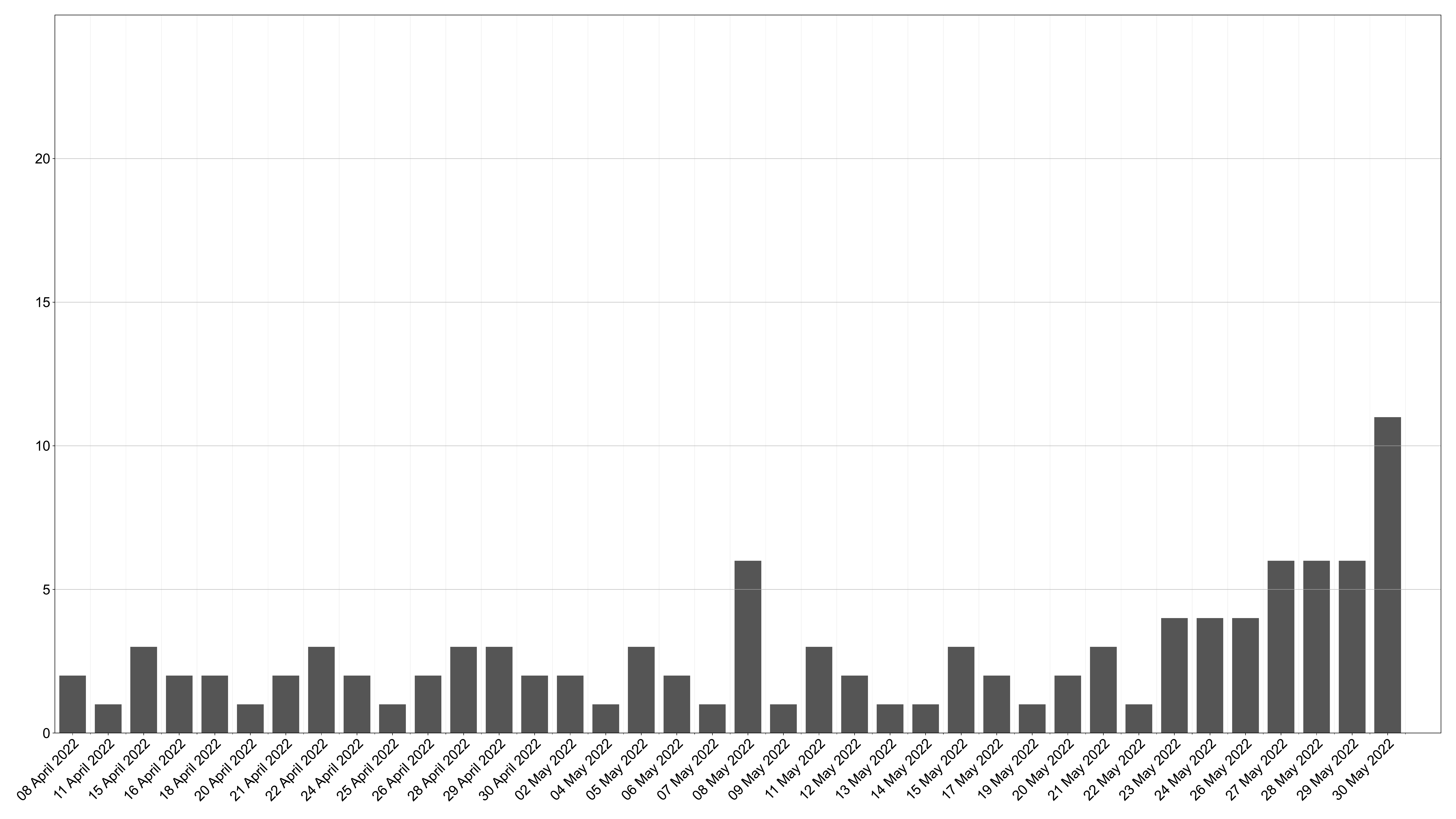}
    \caption{Daily submissions for the Category Detection track (Track 2).}
    \label{fig:cd_submissions}
\end{figure}

\begin{figure}[ht]
    \centering
    \includegraphics[width=0.80\columnwidth]{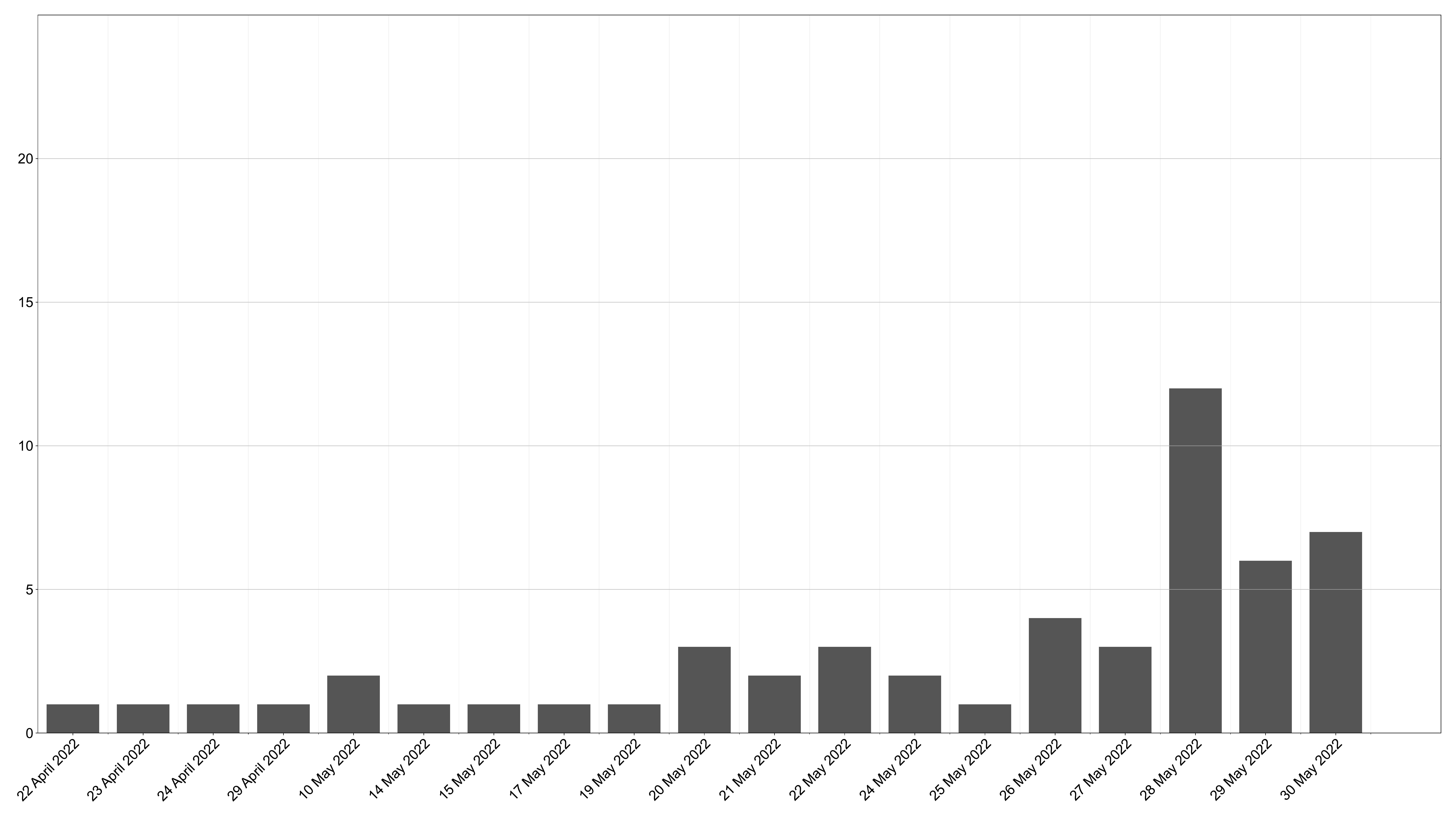}
    \caption{Daily submissions for the Instance Detection track (Track 3).}
    \label{fig:id_submissions}
\end{figure}

The challenge saw an increase in participation over time. The challenge had a final number 360 valid submissions for Track 1 (Instance Classification), 105 submissions for Track 2 (Category Detection), and 53 submissions for Track 3 (Instance Detection). The number of submissions over time is detailed in Figures \ref{fig:ic_submissions}, \ref{fig:cd_submissions}, and \ref{fig:id_submissions}.

\vspace{-5mm}

\FloatBarrier

\bibliographystyle{unsrt}

\bibliography{main}

\end{document}